\documentclass[letterpaper, 10 pt, conference]{ieeeconf}  

\IEEEoverridecommandlockouts                              

\usepackage{graphicx}
\usepackage{amsmath}
\usepackage{amsfonts}
\usepackage{subcaption}
\usepackage{array}

\title{\LARGE \bf
Autonomous 3D Exploration in Large-Scale Environments with Dynamic Obstacles
}

\author{Emil Wiman, Ludvig Widén, Mattias Tiger and Fredrik Heintz
\thanks{This work was partially supported by the Wallenberg AI, Autonomous Systems and Software Program (WASP) funded by the Knut and Alice Wallenberg Foundation.}
\thanks{E. Wiman, L. Widén, M. Tiger and F. Heintz are with the Department of Computer Science,
        Linköping University, Sweden
        {\tt\small emil.wiman@liu.se, ludwi159@student.liu.se, mattias.tiger@liu.se, fredrik.heintz@liu.se}}%
}

\begin{document}

\maketitle
\thispagestyle{empty}
\pagestyle{empty}

\begin{abstract}
Exploration in dynamic and uncertain real-world environments is an open problem in robotics and constitutes a foundational capability of autonomous systems operating in most of the real world. While 3D exploration planning has been extensively studied, the environments are assumed static or only reactive collision avoidance is carried out. We propose a novel approach to not only avoid dynamic obstacles but also include them in the plan itself, to exploit the dynamic environment in the agent's favor. The proposed planner, Dynamic Autonomous Exploration Planner (DAEP), extends AEP to explicitly plan with respect to dynamic obstacles. To thoroughly evaluate exploration planners in such settings we propose a new enhanced benchmark suite with several dynamic environments, including large-scale outdoor environments. DAEP outperform state-of-the-art planners in dynamic and large-scale environments. DAEP is shown to be more effective at both exploration and collision avoidance. 
\end{abstract}

\section{Introduction}

Real world environments change over time. Be it due to construction, renovation, refurbishment, object relocation or deterioration. For robots to function effectively in the real world they must possess the ability to explore their surroundings to build or maintain a 3D world model. Exploration is consequently a foundational capability as it enables the agent to navigate an a priori unknown environment in an effective way and enable the gathering of valuable information about the environment for any number of tasks.

Deliberate exploration is an open problem in robotics. The 3D exploration planning problem is to autonomously explore a potentially large and complex environment as quickly as possible, such that it is covered with a sensor configuration to desired accuracy. The static environment case has been greatly studied for applications such as volumetric exploration \cite{7487281}, surface inspection \cite{surfaceinspec}, object search \cite{8396632}, infrastructure modeling \cite{Yoder2015AutonomousEF}, weed classification \cite{7989676} and 3D reconstruction \cite{8460862} among others. However, most everyday environments of the real world are occupied by people, pets, vehicles and other autonomous agents: The environments are dynamic, not static. Existing techniques do not take into account the presence of dynamic obstacles beyond simple obstacle avoidance behavior \cite{dep}, \cite{9560933}.

\begin{figure}[t]
   \centering
   \includegraphics[scale=0.17]{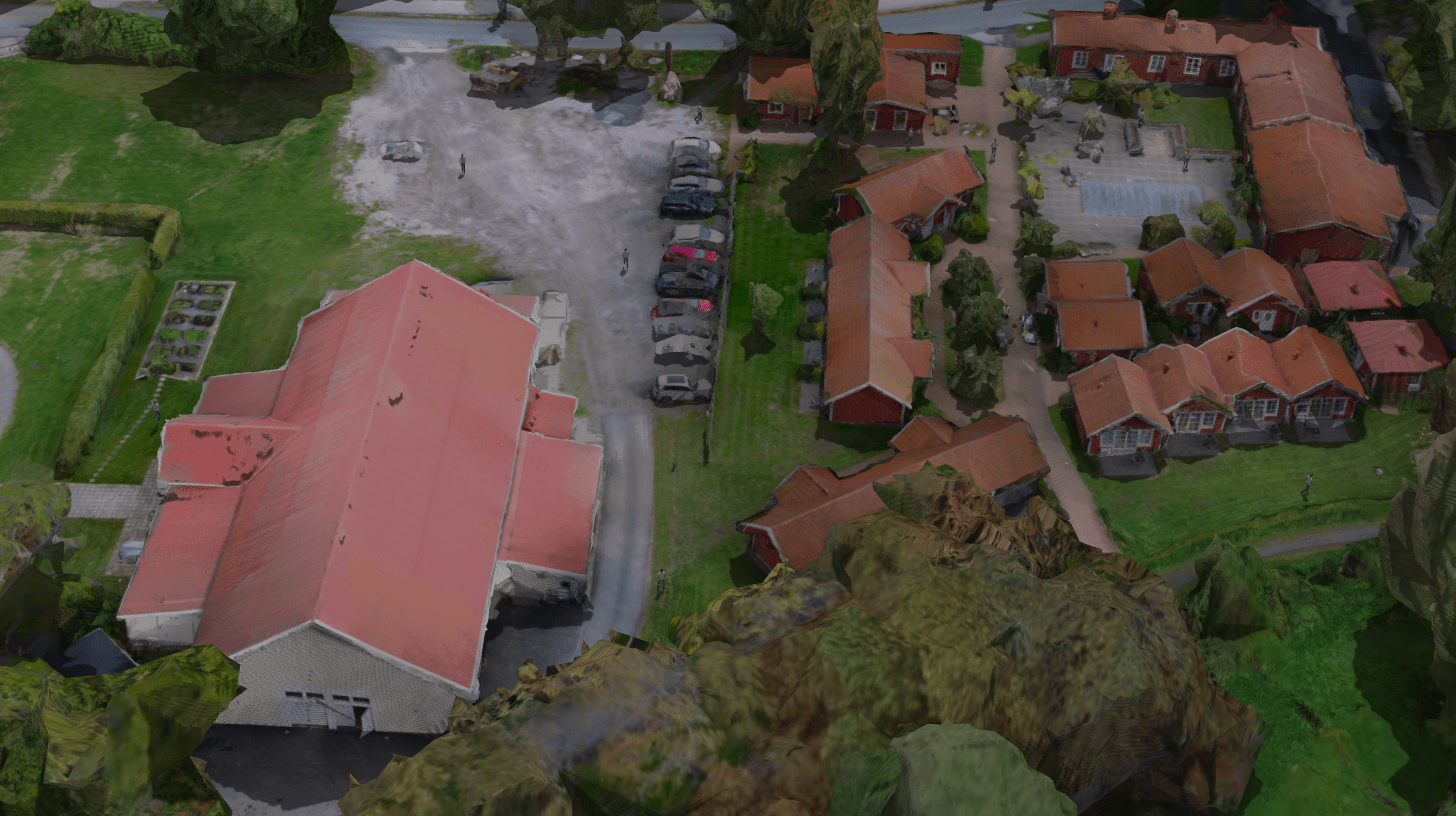}
   \caption{Real world 3D environments can be non-trivial to explore in detail due to large scales, geometrical complexity and the presence of dynamic obstacles wandering about.
   }
   \label{fig: granso10-real}
\end{figure}

Even though it can be possible to force a region to be void of dynamic obstacles, it is often inconvenient and time-consuming. For instance, imagine trying to explore a busy city center like Times Square in New York. The process of removing dynamic obstacles from such a space is both laborious, costly and would cause major annoyance. Furthermore, clearing an area proves especially difficult if the scenario at hand is grand, as in Fig. \ref{fig: granso10-real}. With environments always changing, and busy ones more often than others, it would be far better to be able to effectively explore such environments in the presence of dynamic obstacles. Not to mention if time is of the essence.

We consider the problem of autonomous 3D exploration planning in the large-scale setting (Fig. \ref{fig: granso10-real}) with the presence of dynamic obstacles. Both to avoid dynamic obstacles for safety reasons and to make the exploration more effective by exploiting how the environment change.
The contributions of this work are:

\begin{itemize}
    \item An improved benchmark\footnote{\label{source-code}https://github.com/LudvigWiden/daeplanner} for evaluating exploration planners in environments with dynamic obstacles. It comprise ten maps with varying sizes and complex geometries, reflecting challenging real world environments. Docker is used to provide high reproducibility and compatibility.
    \item A comprehensive evaluation of existing exploration planners utilizing the proposed benchmark. The planners are examined in both a static and dynamic setting to investigate their different strengths and weaknesses.
    \item A proposed planner (DAEP) that demonstrates both \newline superior effectiveness and safety over state-of-the-art.
\end{itemize}

The paper's organization is as follows: In \ref{sec: related-work}, we introduce related research to contextualize the paper's contribution. Defining the dynamic 3D exploration problem takes place in \ref{sec: problem-statement}. The presentation of the proposed method, DAEP, is in \ref{sec: proposed-approach}. The evaluation of this approach is detailed in \ref{sec: experiment-evaluation}. Finally, we provide a summary and conclusions in \ref{sec: summary-and-conclusions}.

\section{Related Work}
\label{sec: related-work}
Autonomous 3D exploration has been under active study for over two decades \cite{613851}, with frontier exploration \cite{613851} as one of the first approaches to tackle 3D exploration. It works by constructing frontiers between explored and unexplored regions of the environment, with the frontiers being explored in some order. Frontier exploration is well established \cite{8206030, 9560933}. A challenge with this kind of approach is how to explore a local neighborhood efficiently and how to take the information gain of the motion between frontier regions into account.

Next, work on the next-best-view (NBV) problem \cite{connolly_1985} from computer vision and computer graphics enabled autonomous NBV exploration planning \cite{banos_2002} which focuses directly on the sensor coverage problem. That is, to find suitable sensor positions to capture the structure of a scene or object. This in turn made efficient local exploration with receding-horizon NBV planning (RH-NBVP) \cite{7487281} possible. It works by combining NVB sampling with rapidly-exploring random trees (RRT) \cite{lavalle_1998} which produce traversable paths between the robot pose and candidate view points. Further, by only executing the first edge provided by the RRT and then repeating the expansion process, the planner becomes adaptive to newly acquired information as it explores the environment.

Autonomous exploration planner (AEP) \cite{aep} combine both paradigms where RH-NBVP is used as local exploration strategy and frontier exploration for global planning. This combination has proved successful, especially in large-scale environments where RH-NBVP may suffer from premature termination. Also, AEP presents a multitude of improvements such as sparse Ray-Casting, dimension reduction of the RRT-sampling space, and the use of Gaussian processes to effectively estimate the potential information gain. The potential information gain translates to the unmapped volume that can be seen from a certain viewpoint and is an important principle we build upon in this work. 

Inspired by the previously mentioned shortcomings, \cite{9507252} presents an online path-planning algorithm for fast exploration and 3D reconstruction of a previously unknown area. It proposes a novel informed sampling-based approach that leverages surface frontiers to sample viewpoints only where high information gain is expected, leading to faster exploration. This approach has been shown to outperform AEP in realistic static exploration scenarios. However, the code is not available and it has consequently not been included in our evaluation.

The first limited steps towards autonomous exploration planning for dynamic obstacles are dynamic frontiers \cite{9560933} and the dynamic exploration planner (DEP) \cite{dep}. The former extends 2D frontier exploration with a new type of frontier that represent one or several dynamic obstacles. These can for example be (detected) people that stand in front of a door opening. Regular frontier exploration would consider the people as part of the map and not assign a frontier region to the occluded door opening. Utilizing the approach of \cite{9560933} these dynamic frontiers will be later explored when people hopefully have moved, and will consequently be able to explore new areas previously blocked by dynamic obstacles.

DEP \cite{dep} instead build a probabilistic roadmap (PRM) \cite{PRM} incrementally, which is used for reactive collision avoidance to find a path around dynamic obstacles when collision are imminent. This way obstacle collisions are potentially reduced, but obstacles have no other consequence for the autonomous exploration itself. DEP denotes this ability to handle dynamic obstacles as a re-plan functionality.

\section{Problem Statement}
\label{sec: problem-statement}
The problem to consider can be formalized as follows. Given a 3D volume
$V \subset \mathbb{R}^3$, the objective of the agent is to explore this volume as completely as possible while avoiding collisions with dynamic obstacles. The volume $V$ consists of two components, namely the free volume $V_{free}(t)$ and the occupied volume $V_{occupied}$. Note that the free volume is subject to temporal change, meaning that at time $t$ might the volume be occupied by a dynamic obstacle. Initially, all poses $\mathbf{p} \in V \subset \mathbb{R}^3$ are unmapped. Thus the objective is to build an internal representation, $M$, that resembles $V_{occupied}$ as closely as possible by exploring the environment. Here $V_{occupied}$ refers to the static environment where dynamic obstacles have been excluded. Moreover, the agent must compute feasible routes that avoid the trajectory of the dynamic obstacles while simultaneously avoiding sub-optimal views in the environment to minimize the exploration time and path length. Due to the highly uncertain and dynamic setting of the environment, this must be solved online.  

\section{Proposed Approach}
\label{sec: proposed-approach}
\begin{figure}[t]
   \centering
   \includegraphics[scale=0.4]{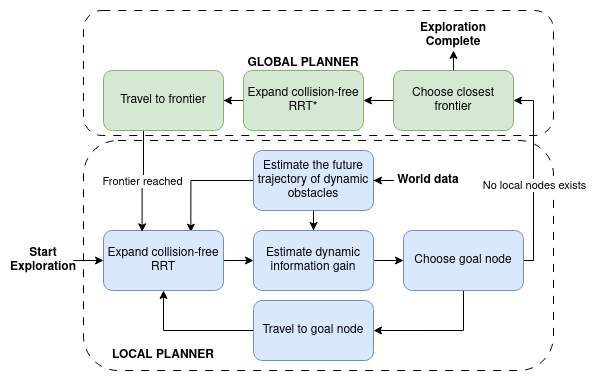}
   \caption{System overview for DAEP. It consists of a local and global planner that collaborate to explore the dynamic environment in an efficient and safe way.}
   \label{fig: system-overview}
\end{figure}

We propose the Dynamic Autonomous Exploration Planner (DAEP), which builds upon AEP and introduce several important modifications and improvements. The strengths of AEP's combined local and global planner is married with a predictor component to ensure collision-free paths. Moreover, the consequences of the temporal presence of dynamic obstacles is considered in the planning itself, allowing DAEP to make deliberate exploration decisions with respect to the dynamic obstacles. See Fig. \ref{fig: system-overview} for an overview of DAEP.

\subsection{Predictor}
To operate in a dynamic environment, a predictor component is needed to estimate the future trajectory of dynamic obstacles. Here, a Kalman filter \cite{kalman} has been employed with a constant velocity motion model. The Kalman filter provides a future distribution of the position of the dynamic obstacle, containing future means and covariances. These can be utilized to handle the uncertainty in the dynamic environment and thus help construct collision-free paths. 

\subsection{Time-based RRTs}
Including a predictor component enables the agent to construct paths that avoid the future trajectories of the dynamic obstacles. This is done by introducing time as a state in the RRT-tree construction, similarly as \cite{6907855}. Each node is assigned a time of arrival, namely the time at which the agent is estimated to reach a certain node. By comparing the time of arrival with the future trajectory for each dynamic obstacle can it be determined whether or not the node is collision-free in the future. This technique has been implemented in the local planner and the global planner.

\subsection{Dynamic Information Gain}
\label{sec:dynamic-information-gain}
\begin{figure}[t]
   \centering
   \includegraphics[scale=0.17]{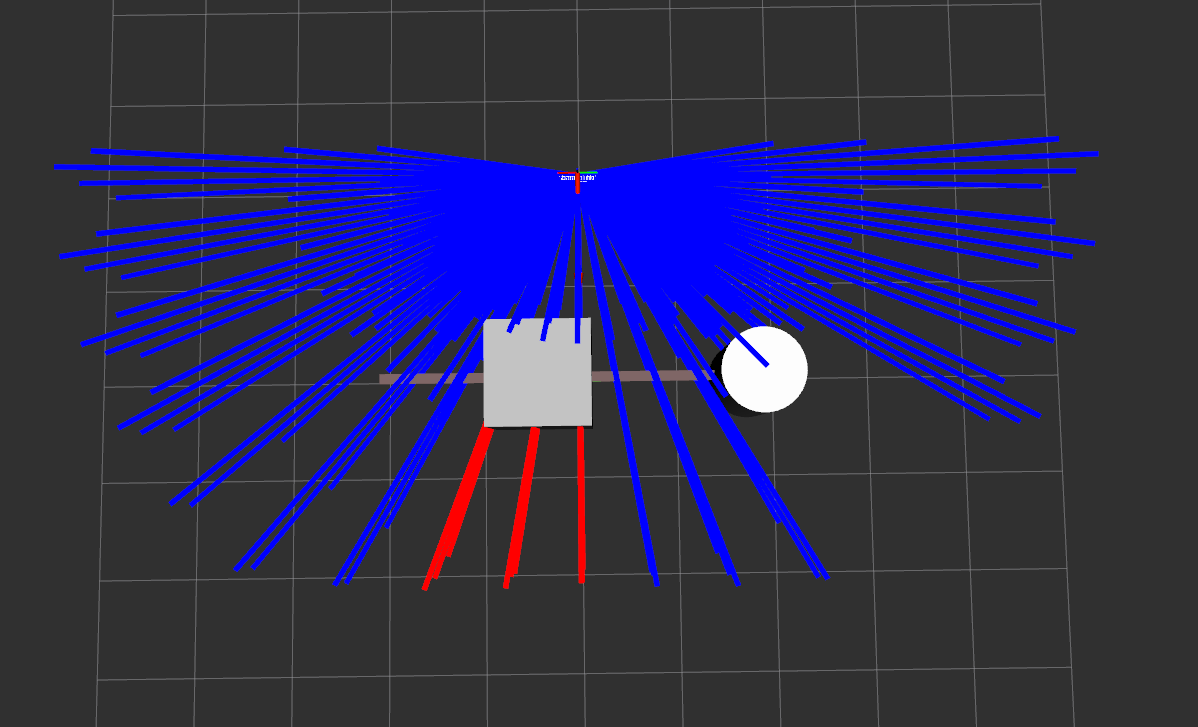}
   \caption{Dynamic information gain. The white circle represents the current position of a dynamic obstacle while the grey square represents its position five seconds into the future. The blue rays represents the field of view of the agent as well as the information gain that can be acquired from that pose. The red rays represents the excluded information gain since it will not be visible upon arrival to the node (top) due to the dynamic obstacle blocking the view.}
   \label{fig: dynamic-gain}
\end{figure}

Due to the dynamic environment is it no longer guaranteed that the estimated potential information gain can be acquired upon arrival to a certain view. This is due to the fact that dynamic obstacles may block the view upon arrival, hence decreasing the information gain acquired. To address these issues has a dynamic score function $s(\textbf{p}, t)$ been introduced (more on this in section \ref{sec:dynamic-score-function}). This function utilizes dynamic information gain $d(\textbf{p},t)$, see Fig. \ref{fig: dynamic-gain}, to produce better decisions in the dynamic environment. Inspecting Fig. \ref{fig: dynamic-gain}, the white circle represents the current position of the dynamic obstacle while the gray square represents its future position according to the predictor component. Arriving at the point where the blue rays originate, the line of sight will be obstructed, resulting in the invisibility of the red rays as a consequence. The dynamic information gain is simply computed as the difference between the blue rays and the red rays (note that the red rays start within the gray square). This dynamic information gain can be estimated during the construction of the RRT and hence assigned to each node.

\subsection{Dynamic Frequency Map}
\label{sec:dynamic-frequency-map}
Dynamic obstacles tend to not navigate uniformly, usually, they follow designated paths or roads. This can be leveraged to enhance decision-making in a dynamic environment. By constructing a heat map of the environment and updating it with the position of dynamic obstacles can a distribution of the historical position of dynamic obstacles be obtained. This Dynamic Frequency Map, $DFM(\textbf{p})$,  can then be utilized to boost areas that have previously shown significant occupancy but are presently unoccupied according to the most recent estimation.

\subsection{Dynamic Score Function}
\label{sec:dynamic-score-function}
To aid the agent in the decision-making process has a new dynamic score function been implemented that extends the score function of AEP with a temporal (see Section \ref{sec:dynamic-information-gain}) and a statistical (see Section \ref{sec:dynamic-frequency-map}) component . The dynamic score in pose $\textbf{p}$ is 

\begin{equation}
    s(\textbf{p},t) = d(\textbf{p},t) \cdot \underbrace{e^{-\lambda \cdot c(\textbf{p})}}_{(0,1]} \cdot (1 + (\zeta \cdot \underbrace{DFM(\textbf{p})}_{[0,1]})
\end{equation}

\noindent where the dynamic score $s(\textbf{p}, t)$ for a specific pose $\textbf{p}$ at time $t$ is determined by the dynamic gain $d(\textbf{p}, t)$ scaled by the cost $c(\textbf{p})$ associated with traveling to that pose. Also, the dynamic scores receive a potential boost $(1 + (\zeta \cdot DFM(\textbf{p})$. Here, $\lambda$ and $\zeta$ are tuning parameters and were manually modified until sufficient behavior was achieved. 

\subsection{Yaw Angle Booster}
During initial experimentation, it was observed that AEP occasionally leaves out certain areas near the map boundary during exploration, leading to large holes in the representation of the environment. This has been noted by \cite{9507252}. The cause of this is that AEP only accumulates volume inside the pre-defined bounding box. This means that when the agent approaches the border of the bounding box, will the volume outside of the box be neglected and hence will the information gain drop drastically leading to more sloppy exploration. This has been addressed in DAEP by boosting the information gain close to the borders artificially. This done by multiplying the computed information gain with some constant $\alpha$, which is given as a parameter. This has improved the exploration close to the borders and secluded the large holes.

\section{Experimental Evaluation}
\label{sec: experiment-evaluation}
To evaluate the performance of DAEP compared to other planners, especially for realistic scenarios with dynamic obstacles, a benchmark has been developed (\ref{sec: benchmark}).
DAEP and three competing planners (RH-NBVP \cite{7487281}, AEP \cite{aep}, DEP \cite{dep}) are evaluated on the benchmark. Due to the code not being available, the planner in \cite{9507252} has not been evaluated. The planners are first evaluated in a number of static environments to gauge their relative performance in the classical case. They are then evaluated in dynamic environments (with dynamic obstacles). Finally, DEP and DAEP are evaluated on large-scale dynamic environments to see how they scale.

\subsection{Benchmark}
\label{sec: benchmark}
The benchmark\footnotemark[1] include ten dynamic scenarios (Table \ref{tab:scenarios}) which can also be run as static worlds. Five of the worlds are from \cite{dep} where we have added difficult dynamic obstacles (people walking) to the previous static worlds \textit{Cafe}, \textit{Maze} and \textit{Apartment}, made them more difficult in \textit{Field}, and kept them as-is in \textit{Auditorium} and \textit{Tunnel}. In the new scenario \textit{Crosswalks} has 4 people crossing back-and-forth, \textit{Patrol} has eight people moving on patrol paths and \textit{Exhibition} have people moving along the walls at a close poster-viewing distance. The large-scale scenario \textit{Village} is a high-res scan of a cotton village with surrounding greenery, with people walking around. There is also an exhibition area and a parking lot.

\begin{table}[t]
    \centering
    \caption{Experimental Parameters}
    \label{tab:exp-params}
    \begin{tabular}{|l|l|l|l|}
    \hline
    \textbf{Parameter} & \textbf{Value} & \textbf{Parameter} & \textbf{Value} \\
    \hline
    Linear Velocity & 0.5 [m/s] & Collision Box [$m^3$] & [0.4, 0.4, 0.1] \\
    Angular Velocity & 1 [rad/s] & Horizontal FoV & 103.2 [deg] \\
    Map Resolution & 0.2 [m] & Vertical FoV & 77.4 [deg] \\
    RRT Ext. range & 1 [m] & Camera Range & 5 [m] \\
    Dyn. Obs. Lin. Vel. & 0.35 [m/s] & Dyn. Obs. Ang. Vel. & 1 [rad/s] \\
    $\lambda$ & 0.75 & $\zeta$ & 0.5 \\
    $\alpha$ & 6 & & \\
    \hline
    \end{tabular}
\end{table}

\begin{table}[t]
  \centering
  \caption{Worlds part of the benchmark.}
  \label{tab:scenarios}
  \begin{tabular}{|l|c|r|r|}
    \hline
    \textbf{World} & \textbf{Origin} & \textbf{Volume [m³]} & \textbf{No. Dyn. Obs.} \\
    \hline
    Cafe & \cite{dep} & 510 & 2 \\
    Maze & \cite{dep} & 865 & 5\\
    Apartment & \cite{dep}  & 1627 & 12\\
    Tunnel & \cite{dep} & 1100 & 2 \\
    Field & \cite{dep} & 1440 & 8\\
    Auditorium & \cite{dep} & 798 & 3\\
    Exhibition & NEW & 450 & 22\\
    Crosswalks & NEW & 450 & 4\\
    Patrol & NEW & 800 & 8\\
    Village & NEW  & 40057 & 15\\
    \hline
  \end{tabular}
\end{table}

\begin{figure}[t]
   \centering
   \includegraphics[scale=0.17]{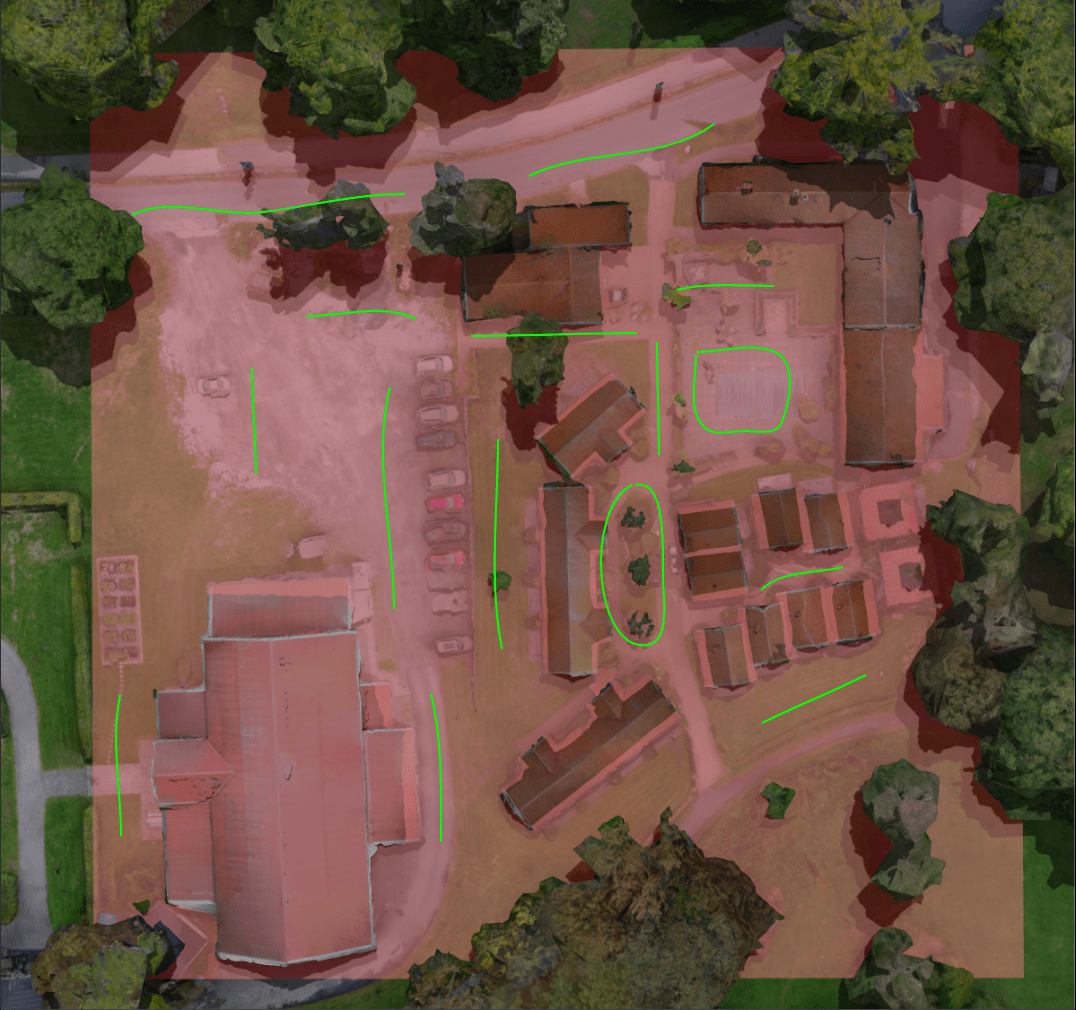}
   \caption{View of the \textit{Village} area. The red bounding box illustrates the volume to be explored. The area is roughly 1 hectare and populated with multiple dynamic obstacles. The paths of the dynamic obstacles are highlighted in green where the lines indicate that the dynamic obstacle move back and forth while a loop indicates that the dynamic obstacles circulate.}
   \label{fig: granso-side-bb}
\end{figure}

The benchmark code itself consists of a Docker solution which simplify getting started and extending the benchmark in the future. It simplifes running all planners on a single machine, despite requirements on different versions of ROS and other conflicting dependencies. Integration of the four planners RH-NBVP, AEP, DEP, and DAEP is provided. The scenarios are simulated with Gazebo 9, using the same simulated quadcopter equipped with a depth camera as in \cite{dep}. The controller supplied by \cite{dep} has been employed in all planners, to avoid alterations of the motion planning. OctoMap \cite{hornung13auro} is used as representation of the internal map. 
The following experiment procedure has been followed:

\begin{itemize}
    \item For each experiment run: A planner, world and mode (with or without dynamic obstacles) is chosen.
    \item The agent starts in one of five  different start locations in the specified world with zero yaw.
    \item The agent travels 1 meter vertically up in the air. A 360-degree rotation is performed to gain initial information about the environment and to ensure free space in the representation to start exploring from.
    \item The exploration algorithm starts and exploration begins.
    \item The exploration continues until the planner signals being finished, or the hard time limit (20 min) is reached.
\end{itemize}

 During experiments the default parameters for each planner\footnotemark[1] has been used. The same experiment parameters are used (Table \ref{tab:exp-params}) unless otherwise specified. Each experiment is repeated five times (i.e. five runs) and the results are reported as $\mu\pm\sigma$ over all specified scenarios' mean run. The different performance measures used are \textbf{C}: Coverage [\%], \textbf{T}: Exploration Time [s], \textbf{PL}: Path Length [m], \textbf{PT}: Planning Time [s] and \textbf{NOC}: Number Of Collisions. Note that the abbreviation DEP refer to DEP with its re-plan functionality enabled while the abbreviation DEP-S constitutes that it is disabled.

\begin{figure*}[t]
	\centering
	\begin{subfigure}[t]{0.32\textwidth}
		\centering
		\includegraphics[width=\linewidth]{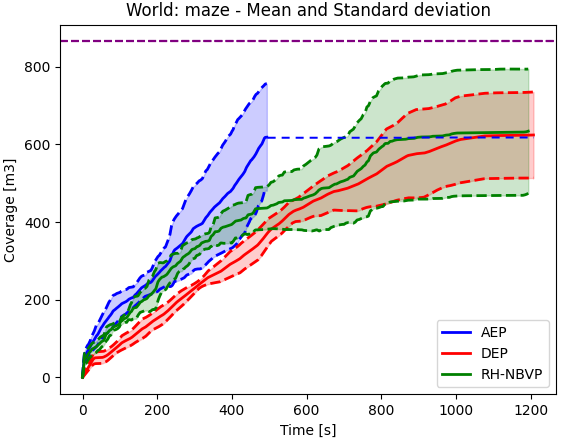}
		\caption{Static world with static planners.}
		\label{fig: maze-exploration}
	\end{subfigure}
    \begin{subfigure}[t]{0.32\textwidth}
		\centering
		\includegraphics[width=\linewidth]{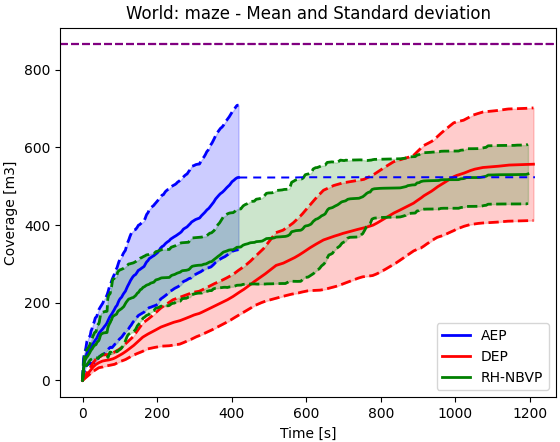}
		\caption{Dynamic world with static planners.}
		\label{fig: maze-exploration2}
	\end{subfigure}
    \begin{subfigure}[t]{0.326\textwidth}
		\centering
		\includegraphics[width=\linewidth]{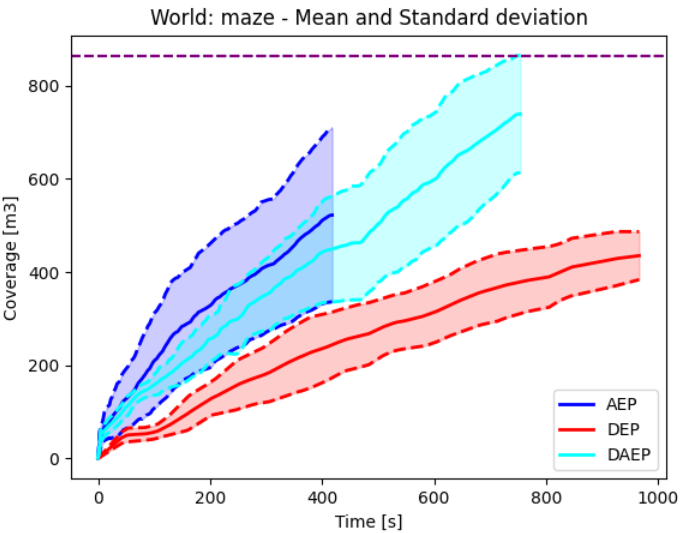}
		\caption{Dynamic world with dynamic planners (AEP as reference).}
		\label{fig: maze-exploration3}
    \end{subfigure}
    \begin{subfigure}[t]{0.32\textwidth}
		\centering
		\includegraphics[width=\linewidth]{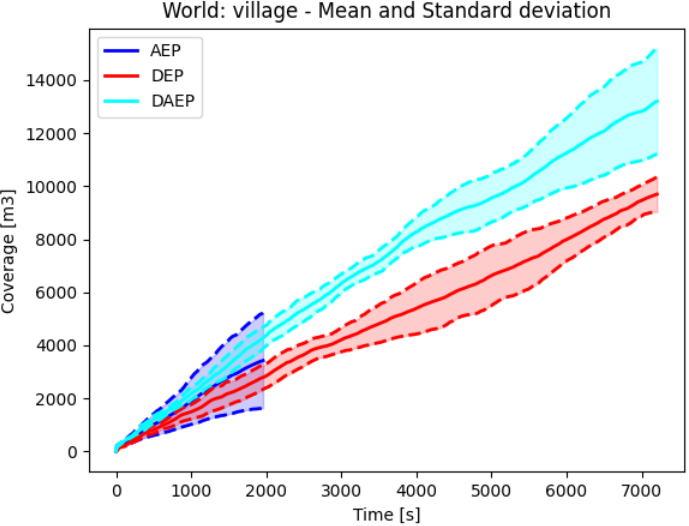}
		\caption{Exploration in the \textit{Village} environment with dynamic obstacles. AEP as reference.}
		\label{fig: granso-avg}
	\end{subfigure}
    \begin{subfigure}[t]{0.32\textwidth}
		\centering
		\includegraphics[width=\linewidth]{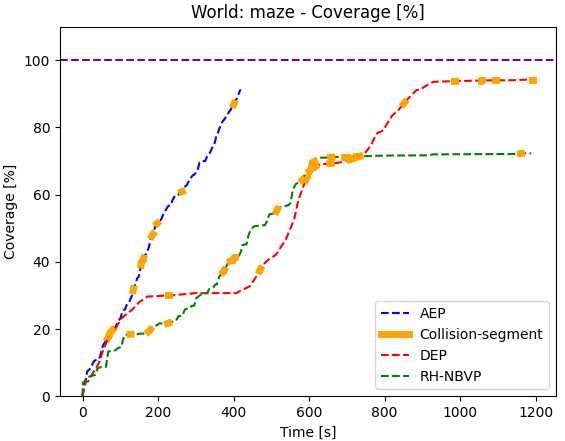}
		\caption{Collisions for static planners.}
		\label{fig: maze-exploration-collisions}
	\end{subfigure}
    \begin{subfigure}[t]{0.32\textwidth}
		\centering
		\includegraphics[width=\linewidth]{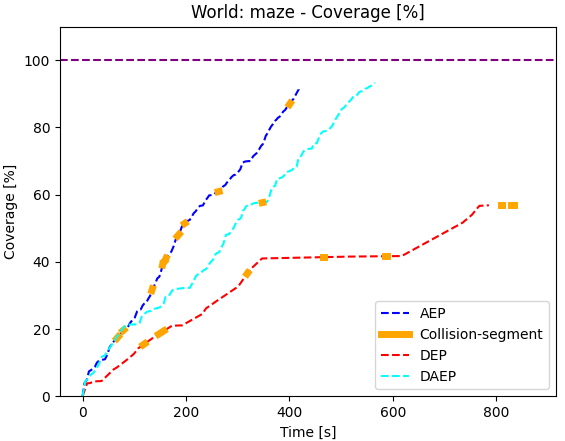}
		\caption{Collisions for dynamic planners.}
		\label{fig: maze-exploration-collisions2}
	\end{subfigure}
    
	\caption{Upper row: Exploration progress in \textit{Maze}. The maximum volume is 865 $m^3$. Here AEP halts the exploration where the envelope stops. Lower row: Exploration progress in the \textit{Village} environment to the left and collision graphs in the middle and to the right for \textit{Maze}. The collision graphs presents the run that accumulated the most coverage. Orange segments indicate collisions and their duration.}
	\label{fig: collected-graphs}
\end{figure*}

\subsection{Static Planners in a Static Environment}
The planners are evaluated on static versions of the first six worlds (same as \cite{dep}) to compare their performance in the classical sense. The aggregated results are shown in Table \ref{tab:agg-res-static-env} and coverage over time is shown for \textit{Maze} in Fig. \ref{fig: maze-exploration}. From Fig. \ref{fig: maze-exploration} it can be observed that AEP manages to explore the environment quicker than both DEP-S and RH-NBVP while acquiring a similar amount of final volume. The same observation is reinforced from Table \ref{tab:agg-res-static-env}, where AEP dominates in terms of exploration time and path length, while DEP manages to find the largest volume on average. All planners face challenges achieving 100\% coverage due to drone size restrictions and imperfect bounding box volume estimates, as evident in upcoming experiments. DAEP was not employed in the initial experiment, as the aim was to assess the performance of alternative planners in a static environment. The purpose of these experiments was to establish a basis for future comparisons and to identify the most suitable planner for future extensions.

\begin{table}[t]
    \centering
    \caption{Results from Different Scenarios and Planners}
    \begin{subtable}{\linewidth}
        \begin{tabular}{| m{0.5cm} | m{2.1cm} | m{2.1cm} | m{2.1cm} |}
        \hline
        & AEP & DEP-S & RH-NBVP \\ \hline
        \textbf{C} & 84.05 $\pm$ 9.53 & \textbf{85.84} $\pm$ 11.28 & 80.55 $\pm$ 12.92 \\ \hline
        \textbf{T} & \textbf{541.14} $\pm$ 401.68 & 1017.35 $\pm$ 248.02 & 1128.76 $\pm$ 168.18 \\ \hline
        \textbf{PL} & \textbf{143.76} $\pm$ 110.25 & 254.81 $\pm$ 70.53 & 211.25 $\pm$ 37.6 \\ \hline
        \textbf{PT} & 55.64 $\pm$ 51.02 & 28.33 $\pm$ 18.39 & \textbf{1.83} $\pm$ 0.33 \\ \hline
        \end{tabular}
        \caption{Results from static environments.}
        \label{tab:agg-res-static-env}
    \end{subtable}
    \begin{subtable}{\linewidth}
        \begin{tabular}{| m{0.5cm} | m{2.1cm} | m{2.1cm} | m{2.1cm} |}
        \hline
        & AEP & DEP-S & RH-NBVP \\ \hline
        \textbf{C} & 80.29 $\pm$ 15.43 & \textbf{84.63} $\pm$ 13 & 77.79 $\pm$ 13.25 \\ \hline
        \textbf{T} & \textbf{460.68} $\pm$ 373.53 & 953.91 $\pm$ 253.45 & 1158.47 $\pm$ 99.91 \\ \hline
        \textbf{PL} & \textbf{111.9} $\pm$ 77.05 & 223.25 $\pm$ 52.8 & 204.82 $\pm$ 30.34 \\ \hline
        \textbf{PT} & 29.95 $\pm$ 24.21 & 25.8 $\pm$ 15.14 & \textbf{2.04} $\pm$ 1.02 \\ \hline
        \textbf{NOC} & \textbf{4.07} $\pm$ 3.83 & 6.47 $\pm$ 5.36 & 6.63 $\pm$ 4.59 \\ \hline
        \end{tabular}
        \caption{Results from dynamic environments with static planners.}
        \label{tab:agg-res-dynamic-env-static-planners}
    \end{subtable}
    \begin{subtable}{\linewidth}
        \begin{tabular}{| m{0.5cm} | m{2.1cm} | m{2.1cm} | m{2.1cm} |}
        \hline
         & AEP & DEP & DAEP \\ \hline
        \textbf{C} & 83.56 $\pm$ 13.58 & 81.85 $\pm$ 18.37 & \textbf{91.25} $\pm$ 6.25 \\ \hline
        \textbf{T} & \textbf{368.78} $\pm$ 335.05 & 775.03 $\pm$ 301.81 & 589.59 $\pm$ 346.44 \\ \hline
        \textbf{PL} & \textbf{91.7} $\pm$ 70.39 & 170.96 $\pm$ 65.15 & 136.43 $\pm$ 66.37 \\ \hline
        \textbf{PT} & \textbf{26.11} $\pm$ 21.33 & 53.7 $\pm$ 39.56 & 79.79 $\pm$ 40.29 \\ \hline
        \textbf{NOC} & 3.36 $\pm$ 3.41 & 6.38 $\pm$ 4.38 & \textbf{0.31} $\pm$ 0.63 \\ \hline
        \end{tabular}
        \caption{Results from dynamic environments with dynamic planners.}
        \label{tab:agg-res-dynamic-env-standard-mp}
    \end{subtable}
    \begin{subtable}{\linewidth}
        \begin{tabular}{| m{0.5cm} | m{2.1cm} | m{2.1cm} | m{2.1cm} |}
        \hline
         & AEP & DEP & DAEP \\ \hline
        \textbf{C} & 8.56 $\pm$ 4.49 & 24.22 $\pm$ 1.66 & \textbf{32.96} $\pm$ 4.98 \\ \hline
        \textbf{T} & \textbf{1094.61} $\pm$ 531.83 & 7175.78 $\pm$ 17.73 & 7201.08 $\pm$ 2.27 \\ \hline
        \textbf{PL} & \textbf{213.03} $\pm$ 104.36 & 893.66 $\pm$ 27.65 & 1280.65 $\pm$ 22.49 \\ \hline
        \textbf{PT} & \textbf{57.08} $\pm$ 30.26 & 2053.47 $\pm$ 163.9 & 930.01 $\pm$ 155.52 \\ \hline
        \textbf{NOC} & 0 $\pm$ 0 & 1.4 $\pm$ 1.0198 & \textbf{0} $\pm$ 0 \\ \hline
        \end{tabular}
        \caption{Results from large-scale dynamic environment \textit{Village}.}
        \label{tab:dynamic-env-large-scale}
    \end{subtable}
    \begin{subtable}{\linewidth}
        \centering
        \begin{tabular}{|l|l|l|}
        \hline
         & DAEP & DEP \\ \hline
        \textbf{C} & \textbf{68.05} & 60.93  \\ \hline
        \textbf{T} & \textbf{31241.5} & 35880.8\\ \hline
        \textbf{PL} & 5132.73 & \textbf{2978.4}  \\ \hline
        \textbf{PT} & \textbf{5763.29} &  20068.1 \\ \hline
        \textbf{NOC} & \textbf{0} & 25 \\ \hline
        \end{tabular}
        \caption{\textit{Village} over 10 hours in a dynamic setting with DAEP and DEP.}
        \label{tab:granso10h}
    \end{subtable}
    \label{tab:combined-results}
\end{table}

\subsection{Static Planners in a Dynamic Environment}
\label{sec: static-planners-in-dynamic-environments}
Next we investigate how AEP, DEP-S, and RH-NBVP are impacted by the presence of dynamic obstacles. The first six worlds are used again, but now filled with dynamic obstacles. The aggregated results can be found in Table \ref{tab:agg-res-dynamic-env-static-planners}. Coverage over time is shown for \textit{Maze} (Fig. \ref{fig: maze-exploration2}) as representative example. Variance increased in all planners (Table \ref{tab:agg-res-dynamic-env-static-planners}), possibly due to dynamic obstacles limiting sight and access to certain areas. Collisions increased during exploration (Fig. \ref{fig: maze-exploration-collisions}), which would be disastrous in real world scenarios. The findings in Fig. \ref{fig: maze-exploration-collisions} are reinforced by Table \ref{tab:agg-res-dynamic-env-static-planners} to occur in general. All planners find less coverage compared to Table \ref{tab:agg-res-static-env}. Similarly for exploration time and planning time, except for the exploration time of RH-NBVP. Finally, each planner collides at least four times on average for each run.

\subsection{Dynamic Planners in a Dynamic Environment}
Introducing the dynamic planners in the dynamic environment should address the issues presented in Section \ref{sec: static-planners-in-dynamic-environments}. Here, DEP and DAEP are employed in the dynamic environment, with AEP as a reference. The experiments have been conducted in the first six worlds, as well as in \textit{Exhibition}, \textit{Crosswalks} and \textit{Patrol}. The aggregated results can be found in Table \ref{tab:agg-res-dynamic-env-standard-mp}. A representative example is shown in Fig. \ref{fig: maze-exploration3} with associated collision rate (Fig. \ref{fig: maze-exploration-collisions2}) for \textit{Maze}. From Fig. \ref{fig: maze-exploration3} it is prominent that DAEP explores the environment faster and considerably more meticulously than DEP. Noticeably, it also explores for a longer period than AEP and thus, manages to gather more volume. Furthermore, AEP and DEP continue to collide frequently, while DAEP collides rarely (e.g. only once in Fig \ref{fig: maze-exploration-collisions2}). The results provided in Table \ref{tab:agg-res-dynamic-env-standard-mp} demonstrate that DAEP manages to accumulate more coverage than DEP and AEP on average. Additionally, it does so with a reduced average exploration time and path length compared to DEP. However, the planning time has increased in DAEP compared to DEP, due to the increased computations needed to handle the dynamic environment. Finally, the number of collisions has decreased significantly for DAEP, compared to DEP and AEP.  

\subsection{Large-Scale Environments}
Finally, we investigate how the planners scale to realistic large-scale outdoor scenarios. Here, we use \textit{Village} which depicts a partial environment of Gränsö castle near the town of Västervik in Sweden, see Fig. \ref{fig: granso-side-bb}. The collected findings for the 2-hour experiment are revealed in Table \ref{tab:dynamic-env-large-scale} and the exploration progress is depicted in Fig. \ref{fig: granso-avg}. Interestingly, it can be observed, from Fig. \ref{fig: granso-avg}, that AEP halts the exploration after only 2000 seconds. Correspondingly, this can be noticed in Table \ref{tab:dynamic-env-large-scale} where AEP collects significantly less coverage. It was found that this is due to a scalability issue in AEP which has been resolved in DAEP. AEP's limited coordinate sampling constrained its exploration range. Moreover, DAEP manages to find a larger amount of volume on average compared to DEP, while completely avoiding collisions.  

\begin{figure}[ht]
   \centering
   \includegraphics[scale=0.15]{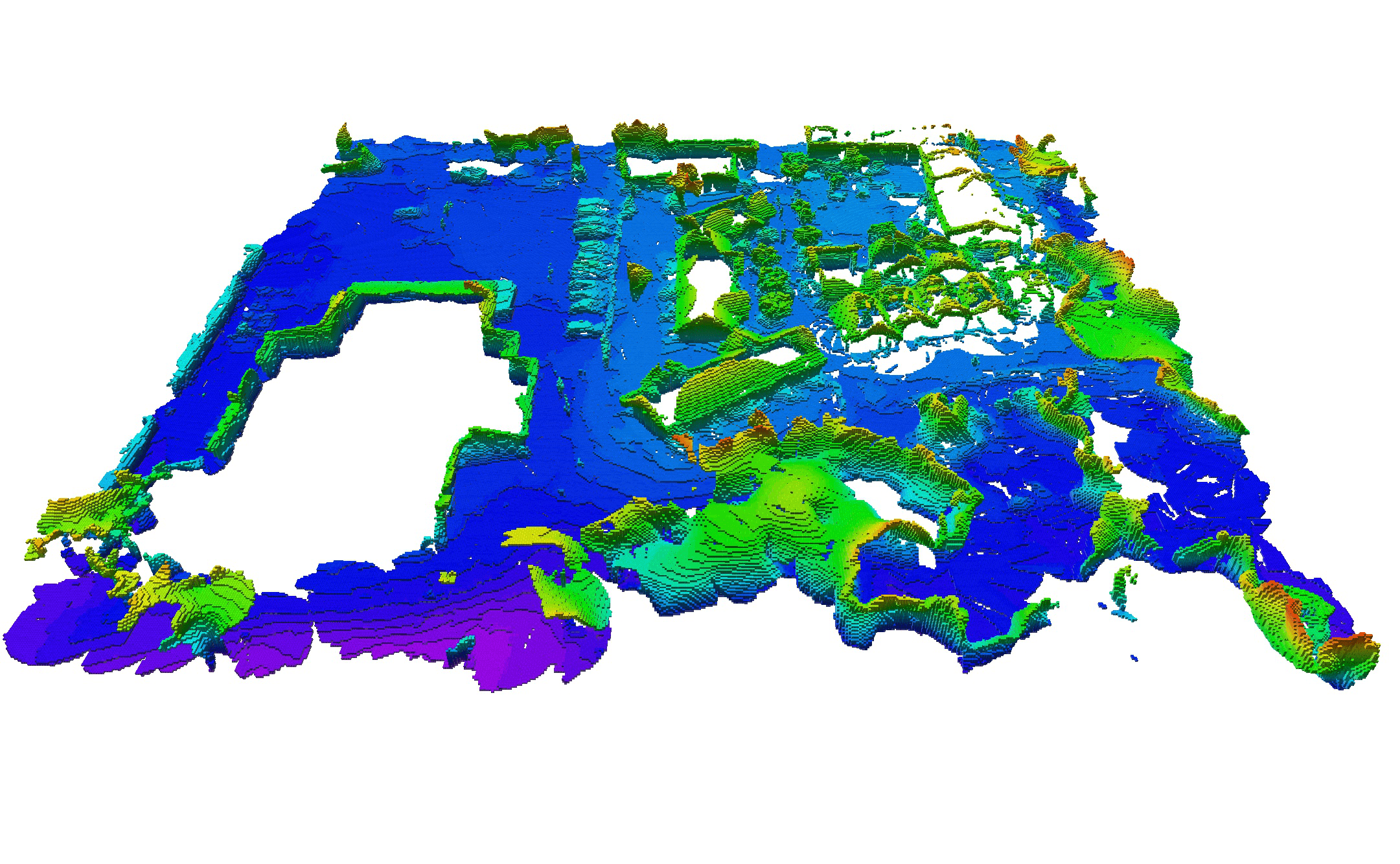}
   \caption{Constructed representation of the \textit{Village} environment from DAEP.}
   \label{fig: granso10-octomap}
\end{figure}

After the 2-hour experiment only roughly 32\% of the \textit{Village} environment was mapped by DAEP. Hence, DAEP and DEP were allowed to continue to explore for a total of 10 hours to push their limits. The results can be found in Table \ref{tab:granso10h} and the corresponding representation of the world is depicted in Fig. \ref{fig: granso10-octomap}. Comparing the real world in Fig. \ref{fig: granso-side-bb} with the representation in Fig. \ref{fig: granso10-octomap}, DAEP is observed to have managed to capture the essential structures and details of the environment. Examining Table \ref{tab:granso10h}, it shows that roughly 68\% of the environment has been mapped after 10 hours while avoiding collisions completely. DAEP outperforms DEP both in terms of coverage but also in number of collisions. Also, observe here that DAEP only plans roughly 18\% of the total exploration time while DEP plans 56\% of the total exploration time.

\section{Summary \& Conclusion}
\label{sec: summary-and-conclusions}
We propose a novel approach to autonomous 3D exploration with dynamic obstacles, DAEP, been presented. DAEP is an extension of AEP with improvements and modifications to handle the presence of dynamic obstacles. A predictor component has been added to facilitate the construction of time-based RRTs. This has in turn been utilized to sample collision-free nodes both in the local and global planner. Furthermore, a novel dynamic score function has been proposed to facilitate safe and efficient navigation in a dynamic environment. Here, the dynamic information gain has been used to predict the potential information gain upon arrival to a new view, while the DFM score has been used to boost areas that have previously been populated. DAEP has shown the ability to outperform both static and dynamic competitors during the experiments. It has also shown the possibility to explore large-scale effectively and safely. In future work, we propose to combine the planner with a more sophisticated motion planner for field tests with real people.

\newpage
\bibliographystyle{IEEEtran}
\bibliography{IEEEabrv, references}

\end{document}